\documentclass[11pt]{article}
\usepackage{coling2020}
\usepackage{times}
\usepackage{url}
\usepackage{latexsym}
\usepackage{booktabs}
\usepackage{multirow}
\usepackage{multicol}
\usepackage{array}
\usepackage{graphicx}
\usepackage{bm}

\colingfinalcopy 

\title{A Corpus for English-Japanese Multimodal Neural Machine Translation with Comparable Sentences
}

\author{Andrew Merritt \\
  Franklin College of Arts and Sciences \\
  University of Georgia \\
  {\tt Andrew.Merritt@uga.edu} \\\And
  Chenhui Chu \\
  Graduate School of Informatics \\
  Kyoto University \\
  {\tt chu@i.kyoto-u.ac.jp}\\\AND
  Yuki Arase \\
  Graduate School of Information Science and Technology \\
  Osaka University \\
  {\tt arase@ist.osaka-u.ac.jp}
  }

\date{}

\begin{document}
\maketitle
\begin{abstract}
Multimodal neural machine translation (NMT) has become an increasingly important area of research over the years because additional modalities, such as image data, can provide more context to textual data. Furthermore, the viability of training multimodal NMT models without a large parallel corpus continues to be investigated due to low availability of parallel sentences with images, particularly for English-Japanese data. However, this void can be filled with comparable sentences that contain bilingual terms and parallel phrases, which are naturally created through media such as social network posts and e-commerce product descriptions. In this paper, we propose a new multimodal English-Japanese corpus with comparable sentences that are compiled from existing image captioning datasets. In addition, we supplement our comparable sentences with a smaller parallel corpus for validation and test purposes. To test the performance of this comparable sentence translation scenario, we train several baseline NMT models with our comparable corpus and evaluate their English-Japanese translation performance. Due to low translation scores in our baseline experiments, we believe that current multimodal NMT models are not designed to effectively utilize comparable sentence data. Despite this, we hope for our corpus to be used to further research into multimodal NMT with comparable sentences.
\end{abstract}

\section{Introduction}
\label{intro}
In recent years, the effectiveness of utilizing image data in tandem with a text corpus to improve the quality of machine translation has been a source of extensive investigation. Several proposals have been made to incorporate visual data, such as using a doubly-attentive decoder for image and text data \cite{calixto_doubly-attentive_2017}, initializing the encoder or decoder hidden state with image features \cite{calixto_incorporating_2017}, and using a deliberation network approach to refine translations using image data \cite{Ive_2019}. However, a common difficulty is the lack of publicly available multimodal corpora, particularly for English-Japanese translation tasks. Currently, two of the only available English-Japanese multimodal datasets are the Japanese extension of the Pascal sentences \cite{funaki_image-mediated_2015} and Flickr30k Entities JP \cite{nakayama-etal-2020-visually}, which is a Japanese translation of the Flickr30k Entities dataset \cite{plummer_flickr30k_2015}.

In order to contribute to the current list of English-Japanese multimodal corpora, we propose a new multimodal English-Japanese corpus with comparable sentences. Comparable sentences are sentences that contain bilingual terms and parallel phrases that describe a similar topic, but are not direct translations \cite{Chu:2015:IPS:2856425.2833089}. This data is of particular interest due to its natural prevalence across various areas of media. For example, e-commerce sites in different countries may have product descriptions for similar products in different languages, or social media users may comment about images in several different languages.

In this study, we created a large comparable training corpus by compiling the existing image captions from the MS-COCO \cite{lin_microsoft_2015} and STAIR \cite{yoshikawa_stair_2017} captioning datasets. 
Furthermore, for validation and testing purposes, we translated a small subset of MS-COCO captions that contain ambiguous verbs. The advantage of comparable sentences in relation to their available quantity can be clearly seen in Table \ref{table:datasets}, with our proposed corpus containing almost twice as many sentence pairs as Flickr30k Entities JP, the current largest parallel multimodal English-Japanese corpus. 
As a benchmark of current multimodal NMT models on our corpus, we performed an English-Japanese translation experiment using several baseline models, which confirmed that current NMT models are not well suited to a comparable translation task.
However, we believe that our corpus can be used to facilitate research into creating multimodal NMT models that can better utilize comparable sentences. 
%
%
\blfootnote{
    %
    %
    \hspace{-0.65cm}  
    %
    %
    This work is licensed under a Creative Commons Attribution 4.0 International License. License details: \url{http://creativecommons.org/licenses/by/4.0/}.
    %
    %
    %
    %
}
\newcommand\T{\rule{0pt}{2.6ex}}  
\newcommand\B{\rule[-1.5ex]{0pt}{0pt}}
\newcolumntype{L}[1]{>{\raggedright\let\newline\\\arraybackslash\hspace{0pt}}m{#1}}
\newcolumntype{C}[1]{>{\centering\let\newline\\\arraybackslash\hspace{0pt}}m{#1}}
\newcolumntype{R}[1]{>{\raggedleft\let\newline\\\arraybackslash\hspace{0pt}}m{#1}}

\begin{table}[t!]
    \centering
    \begin{tabular}{ l | l l r r }
        \toprule
        \T\B Corpus & Source & Type & \T\B \# Images & \T\B \# Sentence Pairs\\
        \hline
        \T\B Pascal Sentences JP  & Pascal Sentences & Parallel  & $1,000$ & $5,000$ \\
        \hline
        \T\B Flickr30k Entities JP & Flickr30k Entities & Parallel & $31,783$ & $63,566$ \\
        \hline
        \T\B \textbf{Our Corpus} & \T\B \textbf{MS-COCO/STAIR} & \textbf{Comparable}  & \T\B $\bm{123,287}$ & \T\B $\bm{123,287}$ \\
        \bottomrule
    \end{tabular}
    \caption{\label{table:datasets}Summary of currently available multimodal English-Japanese corpora compared to ours.}
\end{table}

\section{Dataset Construction}
\begin{figure}
  \includegraphics[width=\linewidth]{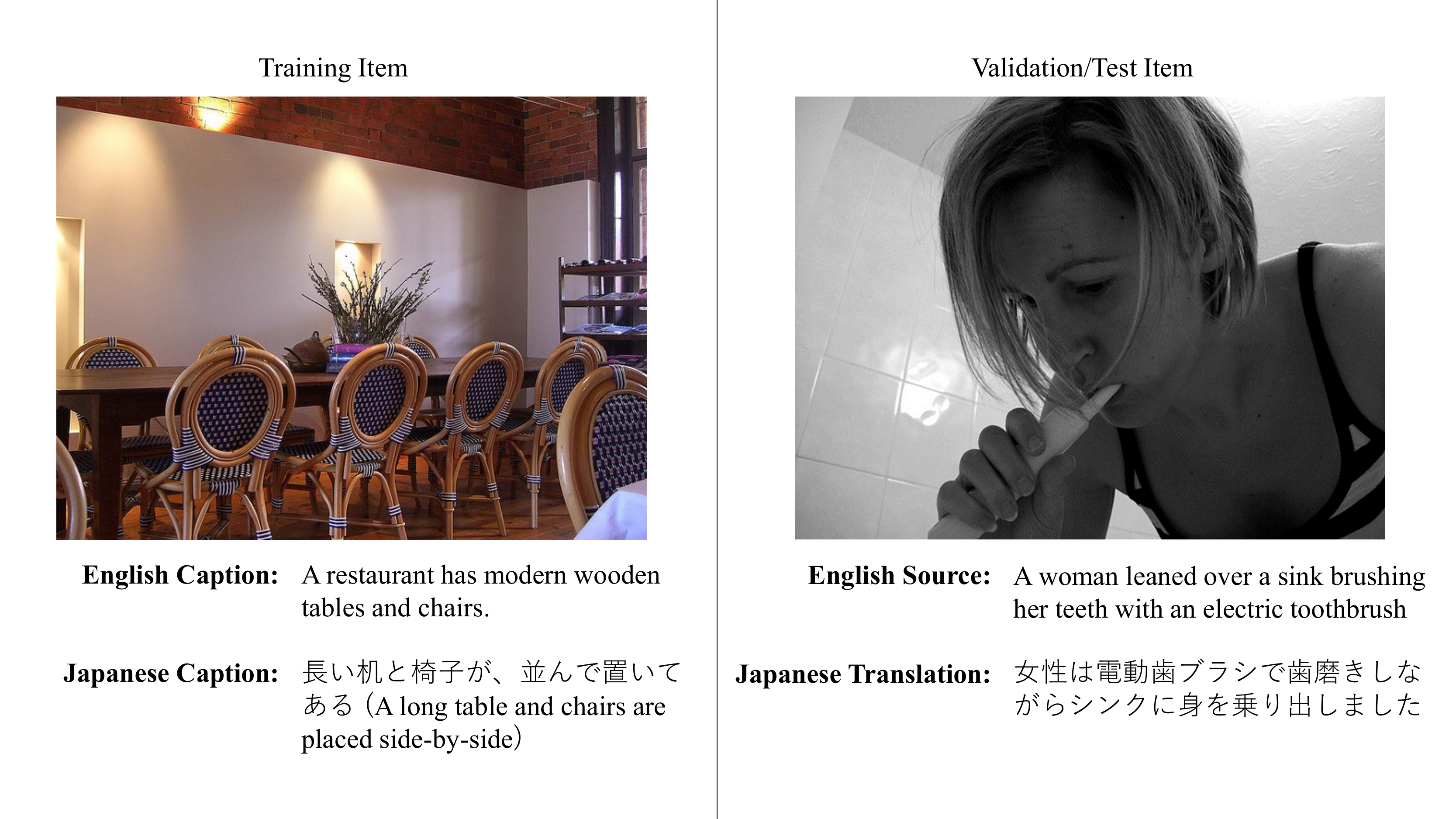}
  \caption{Example of a comparable training item (English translation is in parenthesis) alongside a parallel test item.}
  \label{fig:TrainTestExample}
\end{figure}

Our corpus is an extension of the MS-COCO image captioning dataset \cite{lin_microsoft_2015} and the STAIR Japanese captions for MS-COCO images \cite{yoshikawa_stair_2017}. MS-COCO contains $164,062$ images and $5$ corresponding English captions per image. Conversely, STAIR provides $5$ Japanese captions for each of the images from the 2014 MS-COCO dataset. Because the STAIR Japanese captions were generated independently from the MS-COCO English captions but describe the same images, they are considered comparable. The textual part of our proposed corpus is split into two main portions: the training data, which is comprised of only comparable sentence pairs, and the validation and test data, which is comprised of a relatively small number of parallel sentences. On the left in Figure \ref{fig:TrainTestExample}, there is an example of an English-Japanese comparable sentence pair describing an image of a restaurant. On the right, we give an example of a parallel translation from our test set that describes an image of a woman brushing her teeth. 

\subsection{Training Data}
We started by combining the MS-COCO and STAIR training and validation datasets into one large training set, excluding images and captions that belong to our validation and test data. This left $122,826$ images with $5$ English and Japanese captions per image. Because there are $5$ captions in both languages, there are $25$ possible combinations of sentence pairs that could be associated with a single image. However, for our experiments, we selected one English and Japanese caption in order to create a single comparable sentence pair for each of the images. Overall, the training data totals to $122,826$ images and an equal number of English-Japanese comparable sentence pairs. 

\subsection{Validation and Test Data}
In order to create a test scenario to accompany our training data, we examined a subset of the MS-COCO captions known as Ambiguous COCO \cite{elliott_findings_2017}, which consists of $461$ captions that were chosen due to the ambiguous nature of their verbs. Ambiguous COCO originally only contained English, German, and French captions, so we had the English captions translated into Japanese by a translation company. This company was provided with the English captions as well as the corresponding images, and they were instructed to refer to those images during translation. The translated data was split in half, allocating $230$ images and sentence pairs to the validation data and $231$ to the test data.

\section{Benchmarks}
\subsection{Baseline Models}
To get preliminary performance numbers for our corpus, we tested the translation quality of several baseline NMT models\footnote{The baseline models were implemented with Pytorch 1.0.0 in the OpenNMT-py framework and are available here: \url{https://github.com/iacercalixto/MultimodalNMT}} that we will refer to as TEXT, DAD, and IMG\textsubscript{D}. The TEXT model is an attention-based text-only NMT model \cite{calixto_incorporating_2017} based on \newcite{bahdanau2014neural}. The DAD model is an extension of the TEXT model that incorporates visual features through a doubly-attentive decoder RNN with two separate attention mechanisms for visual features and source sentences \cite{calixto_doubly-attentive_2017}. Finally, the IMG\textsubscript{D} model is another expansion of the TEXT model that incorporates visual features by using them as an input to initialize the hidden state of the decoder RNN \cite{calixto_incorporating_2017}. 
We consider these models to be good baselines for our corpus because the TEXT model is a de facto standard sequence-to-sequence model for text-based NMT, and the DAD and IMG\textsubscript{D} models are representative models for multimodal NMT.

\subsection{Sentence Weighting}
We also investigated the effects of a method to better utilize the comparable sentences in our corpus. 
Inspired by the work of \newcite{Chen_2019}, we implemented a mechanism for weighting sentences during training. The primary goal of this mechanism is to encourage the models to favor sentences that are more closely related to each other during training.
\par To create weights for the training sentence pairs, we started by producing a GIZA++ word alignment\footnote{GIZA++ word alignment was created with mgiza: \url{https://github.com/moses-smt/mgiza}} \cite{brown-etal-1993-mathematics} for the concatenation of our training dataset and a large web-crawled English-Japanese parallel corpus called jParaCrawl \cite{morishita-etal-2020-jparacrawl}. The inclusion of a large parallel corpus helps produce better word alignments. Once the alignments were generated, we averaged the translation probabilities of every word alignment in each sentence by using the bilingual lexicon produced by GIZA++. We then normalized the resulting weights between $0$ and $1$ with min-max normalization. During training, we multiplied the weight for each sentence pair by the corresponding sentence-level loss. The loss function for our sentence weighting is described in the following equation, with $N$ representing the minibatch size, $u\textsuperscript{(n)}$ representing the sentence weight, $T\textsubscript{y}$ representing the last token in the target sentence, $x$ representing the source sentence, $y$ representing the target sentence tokens, and $\theta$ representing the parameters of the neural network:
\begin{equation}L\textsubscript{weighted}=\frac{1}{N} \sum_{n=1}^{N} u^{(n)} \sum_{i}^{T_{y}} \log \left(p\left(y_{i}^{(n)} \mid p\left(y_{<i}^{(n)}, x^{(n)}, \theta\right)\right)\right.\end{equation}

\subsection{Experiment Settings}
First, the visual features were extracted by feeding the images into a pre-trained VGG19 network\footnote{Following \newcite{calixto_incorporating_2017}, we use the activations of the FC7 layer of the VGG19 network in \newcite{simonyan2014deep}'s paper, which encodes information about the whole image} \cite{simonyan2014deep}. Following \newcite{calixto_incorporating_2017}, all of the baseline models were then trained using stochastic gradient descent with Adadelta \cite{zeiler2012adadelta} and minibatches of size $40$. Each model's BLEU \cite{papineni_bleu:_2001} performance was evaluated after every epoch with our validation data, and training was stopped if performance did not improve for $20$ epochs. Once the models finished training, the $230$ test sentences were translated from English to Japanese. We then evaluated the translation quality of the baseline models with and without sentence weighting using the BLEU-4 \cite{papineni_bleu:_2001} and RIBES \cite{isozaki-etal-2010-automatic} evaluation metrics. 

\subsection{Results}
Table \ref{table:results} summarizes the BLEU-4 and RIBES performance of the baseline models with and without sentence weighting. 
Because our training data consists of comparable sentences instead of parallel sentences, it is unsurprising to see the low performance on our corpus. 
In the test scenario without sentence weighting, the IMG\textsubscript{D} model performs the best out of the three models, achieving a BLEU-4 score of $7.32$ and a RIBES score of $0.503$. 
Sentence weighting did show slight BLEU-4 score improvements for both the TEXT and DAD models, but performance was still very low for all models. Conversely, the RIBES scores decreased slightly for all three models after sentence weighting. Overall, the experimental results showed that more complex changes will need to be made to existing models in order to effectively use comparable sentences for training multimodal NMT models.

\newcommand{\ra}[1]{\renewcommand{\arraystretch}{#1}}
\begin{table*}\centering
\ra{1.3}
\begin{tabular}{@{}lrrrcrr@{}}\toprule
& \multicolumn{2}{c}{No Sentence Weighting} & \phantom{ab}& \multicolumn{2}{c}{With Sentence Weighting} &
\phantom{ab}\\
\cmidrule{2-3} \cmidrule{5-6}
& BLEU-4 & RIBES && BLEU-4 & RIBES\\ \midrule
TEXT & $6.57$ & $0.501$ && $7.13$ & $0.493$\\
IMG\textsubscript{D} & $7.32$ & $0.503$ && $7.25$ & $0.491$\\
DAD & $7.22$ & $0.493$ && $7.44$ & $0.483$\\
\bottomrule
\end{tabular}
\caption{\label{table:results}BLEU-4 and RIBES performance of baseline models trained on our corpus.}
\end{table*}

\section{Conclusion}
In this paper, we have proposed a new multimodal English-Japanese corpus with comparable sentences. Based on the baseline performance of this data, we believe that current multimodal NMT models are not well suited to this type of task, and further research is required in order to better leverage the comparable sentences and images together in order to improve translation performance. In the future, we hope to see our corpus used to encourage research into multimodal machine translation tasks with comparable sentences instead of parallel sentences. 

\section*{Acknowledgements}
\label{sec:ack}
This work was supported by Microsoft Research Asia Collaborative Research Grant and Grant-in-Aid for Young Scientists \#19K20343.

\bibliographystyle{coling}
\bibliography{coling2020}

\end{document}